\documentclass{article}

\usepackage{neurodata}
\usepackage{lipsum}

\usepackage[utf8]{inputenc} 
\usepackage[T1]{fontenc}    
\usepackage{hyperref}       
\usepackage{url}            
\usepackage{booktabs}       
\usepackage{amsfonts}       
\usepackage{nicefrac}       
\usepackage{microtype}      
\usepackage{xcolor}         

\usepackage{multirow}
\usepackage{array}
\usepackage{graphicx}
\usepackage{amssymb}
\usepackage{amsmath}
\usepackage{amsthm}
\usepackage{mathtools}

\usepackage[createShortEnv]{proof-at-the-end}
\newtheorem{theorem}{Theorem}[section]

\newtheorem{corollaryinner}{Corollary}[theorem] 
{
\if\relax\detokenize{#1}\relax
\else
\ifcsname #1-used\endcsname
  \expandafter\xdef\csname #1-used\endcsname{\the\numexpr\csname #1-used\endcsname+1}%
\else
  \expandafter\gdef\csname #1-used\endcsname{1}%
\fi
\renewcommand{\thecorollaryinner}{\ref{#1}.\csname #1-used\endcsname}%
\fi
\corollaryinner
}
{\endcorollaryinner}
\newtheorem{definition}{Definition}[section]
\graphicspath{ {images/} }

\title{Polarity is all you need to learn and transfer faster}

\author[1,5]{Qingyang ~Wang\thanks{qwang88@jhu.edu}}
\author[2,5,6]{Michael A.~Powell}
\author[4,5]{Eric ~Bridgeford}
\author[2,5]{Ali ~Geisa}
\author[1,2,3,4,5]{Joshua T.~Vogelstein}
\affil[1]{Department of Neuroscience, Johns Hopkins University}
\affil[2]{Department of Biomedical Engineering, Johns Hopkins University}
\affil[3]{Institute for Computational Medicine, Kavli~Neuroscience~Discovery Institute, Johns Hopkins University}
\affil[4]{Department of Biostatistics, Johns Hopkins University}
\affil[5]{Center for Imaging Science, Johns Hopkins University}
\affil[6]{Current location: United States Military Academy, Department of Mathematical Sciences, West Point NY US}

\begin{document}

\maketitle

\begin{abstract}
    Natural intelligences (NIs) thrive in a dynamic world – they learn quickly, sometimes with only a few samples. In contrast, artificial intelligences (AIs) typically learn with a prohibitive number of training samples and computational power. What design principle difference between NI and AI could contribute to such a discrepancy? Here, we investigate the role of weight polarity: development processes initialize NIs with advantageous polarity configurations; as NIs grow and learn, synapse magnitudes update, yet polarities are largely kept unchanged.  We demonstrate with simulation and image classification tasks that if weight polarities are \textbf{adequately} set \textbf{\textit{a priori}}, then networks learn with less time and data. We also explicitly illustrate situations in which \textit{a priori} setting the weight polarities is disadvantageous for networks. Our work illustrates the value of weight polarities from the perspective of statistical and computational efficiency during learning.
\end{abstract}

\section{Introduction}\label{sec:intro}
Natural intelligences (NIs), including those of animals and humans, are able to learn and adapt rapidly in real world environments with limited samples. Artificial intelligences (AIs), specifically deep neural networks (DNNs), can now compete with or even surpass humans in certain tasks, e.g.,  the game GO \citep{Silver2017}, object recognition \citep{Russakovsky2015}, protein folding analysis \citep{Jumper2021}, etc. However, a DNN is only capable of such achievements when prohibitive amounts of data and training resources are available. Such gaps in learning speed and data efficiency between NI and AI have baffled and motivated many AI researchers. A subfield of AI is dedicated to achieving few-shot learning using DNNs \citep{Hoffer2014, Spoel2015, Vinyals2016, Snell2017, Finn2017}. Many research teams have achieved high performance on benchmark datasets \citep{Lazarou2022, Bendou2022a}. However, the products of such engineering efforts greatly deviate from the brain. What are the design principle differences between NIs and AIs that contribute to such a gap in learning efficiency? In this paper, we propose two possible design differences under one theme: weight polarity. 

NIs are blessed with hundreds of millions of years of optimization through evolution. Through trial and error, the most survival-advantageous circuit configurations emerge, refine, and slowly come into the form that can thrive in an ever-changing world. Such circuit configurations are embedded into genetic code, establishing a blueprint to be carried out by development. It is hypothesized that such canonical circuits provide an innate mechanism that facilitates rapid learning for NIs \citep{Zador2019}. Unlike transfer learning for AIs where the entirety of the weights is adopted as a form of prior knowledge, NIs pay substantial, if not exclusive, attention to polarity patterns (excluding magnitudes) when transferring such canonical circuits across generations. On the one hand, polarity is a more compressed knowledge carrier than weight, rendering it a genetic-code-saving design choice. On the other hand, is polarity alone enough to transfer knowledge between networks? NIs suggest yes; AIs' answers are yet to be explored. 

Furthermore, post-development neuronal connections in the brain rarely see polarity switch \citep{Spitzer2017}. After development, NIs learn and adapt through synaptic plasticity -- a connection between a pair of neurons can change its strength but rarely its excitatory or inhibitory nature; on the contrary, a connection (weight) between a pair of units in a DNN can freely change its sign (polarity). For the rare times such phenomenon have been observed, they never appeared in sensory and motor cortices \citep{Spitzer2017} where visual, auditory and motor processing take place. It seems a rather rigid design choice to fix a network's connection polarity. Is it a mere outcome of an implementation-level constraint? Perhaps it is quite difficult for synapses to switch polarities. Or could it be that fixing polarity is a more efficient learning strategy? This is yet another unexplored design principle AIs may borrow from NIs.  

This paper provides some thoughts and evidence in applying these two design principles to DNNs. We first discuss the trade-off between representation capacity and learning speed when weight polarity is fixed for networks (Section~\ref{sec:adv}). We experimentally show that if the weight polarities are \textbf{adequately} set \textbf{\textit{a priori}}, then networks can learn with less time and data (simulated task (Section~\ref{sec:adv}) + two image classification tasks (Section~\ref{sec:cv})). We also discuss how the quality of the polarity configuration affects a DNN's learning efficiency (Section~\ref{sec:adv}-\ref{sec:cv}). We further find transferring + fixing polarities is even superior to transferring weights (Section~\ref{sec:finetune}). Our results point to an unexplored direction in the machine learning community: polarity, not weight, may be the more effective and compressed medium for transferring knowledge between networks. To complete our discussion, we further discuss what we may lose when weight polarities are set \textit{a priori} (Section~\ref{sec:disadv}). 

By discussing both the advantages and disadvantages of setting polarity \textit{a priori}, we provide some insights on how to make AI more statistically and computationally efficient. 

\section{What do we gain by setting weight polarity \textit{a priori}?}\label{sec:adv}
Networks need both positive and negative weights to function \citep{wang2023} - a DNN with all non-negative weights is not a universal approximator. Even when both polarities are present in a network, constraining a network’s weight polarity pattern limits its representation capacity: when only half of the range is available to each connection, the reduction in total possible network patterns grows exponentially with more edges in the network. It seems counter-intuitive for any network to have willingly chosen to give up on a vast portion of its representation capacity. Are they gaining elsewhere? Our thought is: maybe they learn faster. We write down such representation capacity and speed trade-off concisely into Lemma~\ref{lma4} and prove it in a constrained setting. We provide experimental evidence afterwards.


\begin{lemmaE}[capacity-speed trade-off][end, restate]
\label{lma4}
If the weight polarities are set \textit{a priori}, such that the function is still representable, then the network can learn faster. 
\end{lemmaE}
\vspace{-2mm}
We prove Lemma~\ref{lma4} for single-hidden-layer networks, with the following assumptions:
\vspace{-2mm}
\begin{enumerate}
    \item[] Assumption 1: The weights take on discrete values. This is essentially true for all DNNs implemented on silicon chips where all continuous variables are discretized;
    \item[] Assumption 2: Exhaustive search is the learning algorithm. 
\end{enumerate}
\vspace{-2mm}
\begin{proofE}
A feedforward DNN can be described as a graph \\
\begin{equation*}
    G=(V,E), w : E \rightarrow \mathbb{D}.
\end{equation*}
Nodes of the graph correspond to neurons (units), where each neuron is a function $\sigma^{(l)}_j(x) = \sigma(W_j^{(l)}x + b^{(l)}), j\in[n_l]$. All the weights for the network take on values from the set $\mathbb{D} = \{-d, \dots, 0, \dots, d\}$ for some $d \in \mathbb{N}$. The network is organized in \textit{layers}. That is, the set of nodes can be decomposed into a union of (nonempty) disjoint subsets, \(V = \dot\cup_{l=0}^{L}V_l\), such that every edge in $E$ connects some node in $V_{l-1}$ to some node in $V_l$, for some \(l\in[L]\). Assume we have fully connected layers. Then the number of incoming edges per node is $|V_{l-1}|$. Let $|V_0|$ be the input space dimensionality. All nodes in $V_0$ (input layer) and $V_L$ (output layer) are distinct. \\
Let $|G|$ denote the total number of distinct weight patterns of the graph $G$. Then for a single hidden layer network where $L=2$, we have:
\begin{equation}
    |G| = |\mathbb{D}|^{|E|} = |\mathbb{D}|^{(|V_0|*|V_1| + |V_1|*|V_2|)}
\end{equation}
Then
\begin{equation}
    \frac{|G|}{|G_{polarityFrozen}|} = {(\frac{2d+1}{d+1})}^{(|V_0|*|V_1| + |V_1|*|V_2|)}
\end{equation}
Assume a different weight pattern represents a different function (trivially holds in linear + full rank weight case), then for every single representable function, there always exists a set of weight polarity configurations $G_{correct}$ such that ${G_{correct}}\subseteq{G_{polarityFrozen}}$, therefore 
\begin{equation}
    \frac{|G_{correct}|}{|G_{\text{polarityFrozen}}|} \ll \frac{|G_{correct}|}{|G|}
\end{equation}
This means setting weight polarity \textit{a priori} in an adequate way constraints the combinatorial search space to have much higher proportion of correct solutions, hence easier to learn under exhaustive search algorithm (Lemma~\ref{lma4}).
\end{proofE}

Next, we use simulation to show that when trained with Freeze-SGD (Algorithm~\ref{algo}), networks indeed learn more quickly when polarities are set \textit{a priori} in such a way that the function is still representable (Def~\ref{def:representable}). 
\begin{definition}[representability]\label{def:representable}
With input space $\mathcal{X}$, we say a function $f$ is representable by a DNN $F$ when $\forall x \in \mathcal{X}, \epsilon>0, \; \exists F, \text{such that} |f(x)-F(x)|<\epsilon$.  
\end{definition}
\paragraph{Freeze-SGD} We design our freeze training procedure to be exactly the same as SGD (Adam optimizer) except for one single step: after each batch, all weights are compared to the preset polarity template, and if any weight has switched polarity in the last batch, it is reverted back to the desired polarity/sign (see Algorithm~\ref{algo}). As our goal is to see the pure effect of fixing weight polarity, we did not adopt any bio-plausible learning algorithms as they may introduce confounding factors.

We compared four training procedures in general: 
\begin{enumerate}
    \item \textbf{Frozen-Net Sufficient-Polarity}: Weight polarities were set \textit{a priori}. The polarity pattern was chosen based on a rule that ensures the configuration is adequate for learning the task, i.e., the polarity pattern carries expert knowledge about the data. 
    \item \textbf{Frozen-Net RAND-Polarity}: Weight polarities were set \textit{a priori} randomly: $Bernoulli(0.5)$. The polarity pattern does not carry any prior information about the data.
    \item \textbf{Fluid-Net RAND-Polarity}: Weights (containing polarities) were initialized randomly; weight polarities were free to change throughout the training procedure. 
    \item \textbf{Fluid-Net Sufficient-Polarity} Weight polarities (not magnitudes) were initialized with prior knowledge; weight polarities were free to change throughout the training procedure. This scenario will only be discussed in Section~\ref{sec:cv} on image classification tasks. 
\end{enumerate}

\paragraph{Controlled Experiments} To see the pure effect of setting weight polarity \textit{a priori}, we require a suite of controlled measures. We controlled the following factors across all conditions: the weight magnitude distribution (Supplementary Figure~\ref{fig:mag_dist}), network architecture, training samples (n varies, see Figure 1), learning rate, batch sequence of the training data, and the validation samples across all scenarios. More details about the experiments can be found in the Appendix Section~\ref{sec:methods}. 

\subsection{XOR-5D Simulation Results}\label{sec:5D}
\begin{figure}[!ht]
\includegraphics[width=\textwidth]{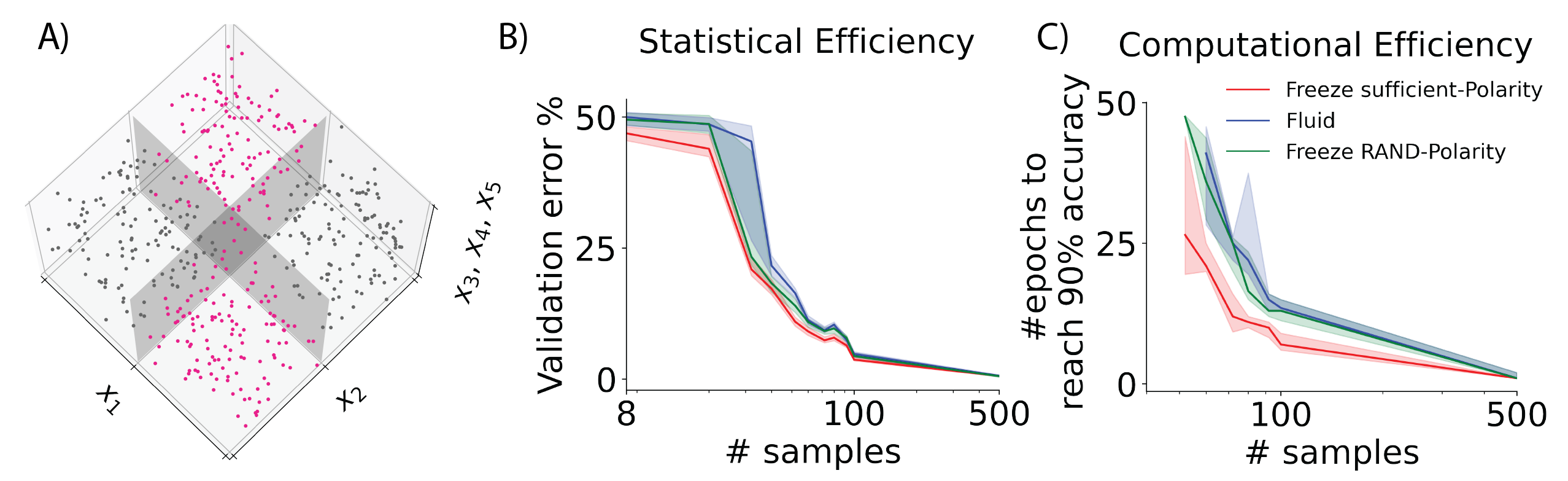}
\caption{\textbf{Adequately setting weight polarity \textit{a priori} allows networks learn faster (fewer epochs) and with less data (fewer training samples).} Single hidden layer networks with 64 hidden units were trained to learn XOR-5D (Section~\ref{sec:5D}). In all scenarios, networks were trained for 100 epochs. A) XOR-5D, a binary classification problem where only the first two dimensions are relevant to the task (XOR), and the other three dimensions are noise following a normal distribution $\mathcal{N}(1,1)$. B) Statistical efficiency: At convergence, adequately configured freeze networks (red) achieve the same level of performance (\% error) with less data; randomly configured freeze networks (green) use similar and oftentimes less data than fluid (blue), e.g., to reach 80\% accuracy (20\% error). C) Computational efficiency: Setting weight polarity in a data-informed way (red) makes networks learn more quickly -- this is true across all data-scarcity levels; when weight polarities are randomly fixed (green), networks learn as fast as their fluid counterparts (blue), and sometimes faster (e.g., training sample sizes = 80 and 92). For all experiments, n=50 trials. All curves correspond to medians with shaded regions 25\textsuperscript{th}-75\textsuperscript{th} percentiles. }
\label{fig:1}
\end{figure}

We used 5-dimensional XOR (XOR-5D) as the binary classification task for our simulation; only the first two dimensions are relevant to the task, and the remaining three dimensions are noise following a normal distribution $\mathcal{N}(1,1)$ (Figure~\ref{fig:1} panel A). For Frozen-Net Sufficient-Polarity, the polarity template is pre-set in this way: for each hidden unit, the polarity of the output edge is the sign product of the first two dimensions’ input weights. 

We first tried four different weight reset methods (details in Algorithm~\ref{algo}) and found they give us similar results and thus did not matter for our primary questions (Supplementary Figure~\ref{fig:sup}). For the rest of this paper, we chose the \textit{posRand} method, whereby we reset weights to a small random number of the correct sign. 

When the polarities are fixed in such a way that it is sufficient to learn the task (red), networks always learn faster than networks without polarity constrains (blue) (Fig~\ref{fig:1} panel B). This advantage is true across all data scarcity levels and is particularly valuable when data is scarce. When only 60 or 72 training samples were available, Frozen-Net Sufficient-Polarity on average takes 58\% and 48\% of the standard Fluid-Net training time, respectively, to reach the same level of accuracy. When weight polarities are randomly chosen (green), networks learn as fast as their Fluid-Net counterparts (blue), and sometimes faster (e.g., training samples = 80 and 92).  

Frozen-Net Sufficient-Polarity not only saves time, but it also requires less data (Figure~\ref{fig:1} panel C). At convergence, Frozen-Net Sufficient-Polarity (red) takes fewer samples to reach the same level of accuracy when compared to the standard Fluid-Net (blue). Randomly configured Frozen-Net (green) uses similar and oftentimes less data than Fluid-Net (blue) (e.g., to reach $80\%$ accuracy, Frozen-Net RAND-Polarity uses less data than Fluid-Net).
\\

We showed that setting weight polarity \textit{a priori} makes networks learn in less time, with less data, provided the function is still representable. Even randomly configured Frozen-Nets show comparable and sometimes better performance than Fluid-Nets. Such a result is striking from an optimization perspective: Frozen-Net is at a disadvantage by design because the weight resetting step in Freeze-SGD (see Algorithm~\ref{algo}) fights against the gradient update, and part of the error update information is lost in this process. Regardless of this disadvantage during optimization, our Frozen-Net Sufficient-Polarity consistently outperforms Fluid-Net; even Frozen-Net RAND-Polarity is never worse than Fluid-Net. Combined, these results show we may be able to help AIs learn quickly and with fewer samples by doing two things: 1) fix weight polarities, and 2) choose the polarity pattern wisely. We will tease apart the effect of these two factors in the next section. 

\section{Effectiveness of setting weight polarity \textit{a priori} in image classification tasks}\label{sec:cv}

\begin{figure}[!ht]
\includegraphics[width=\linewidth]{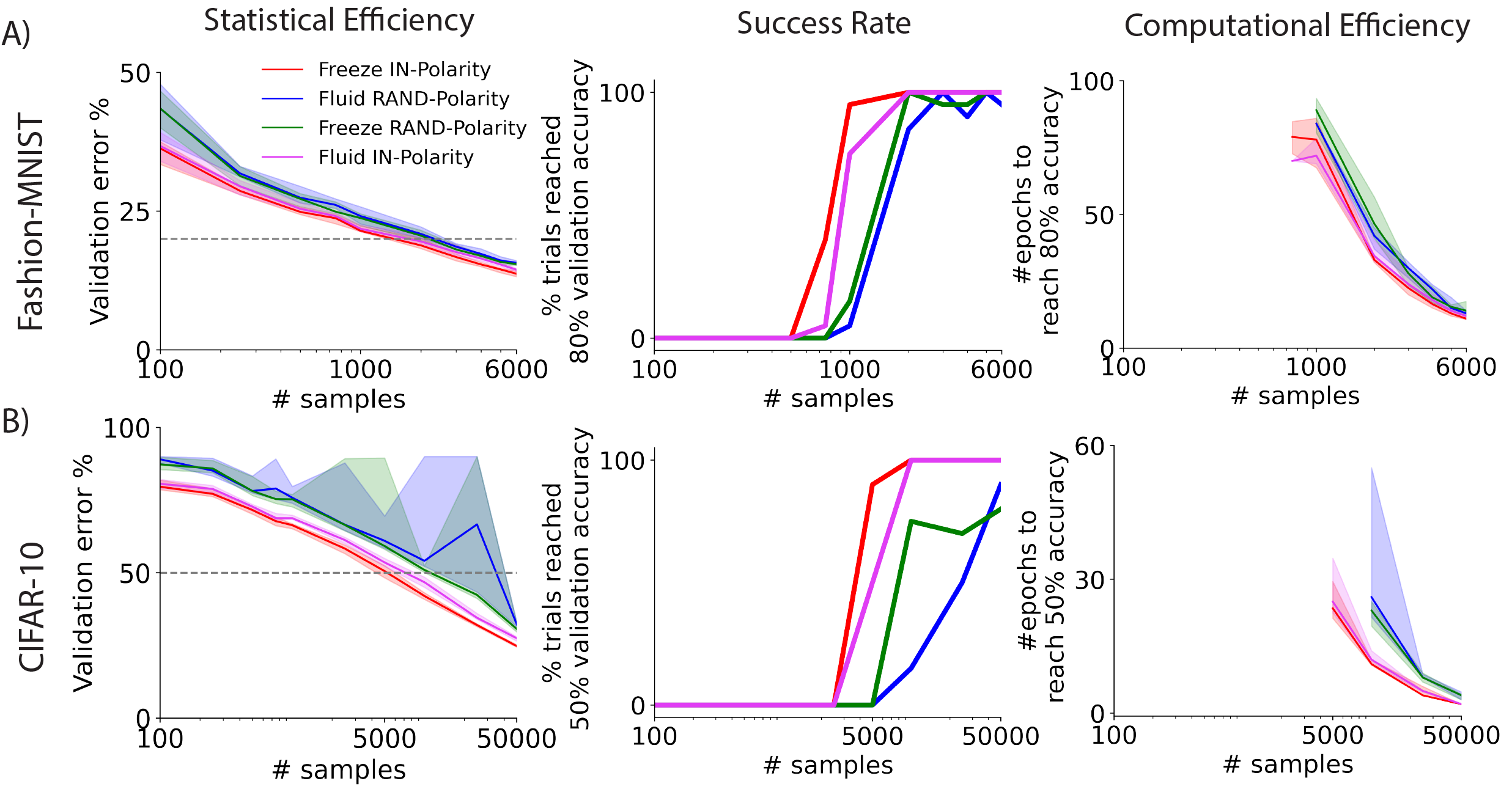}
\caption{\textbf{DNNs with frozen ImageNet-Polarity (IN-Polarity) learn more quickly and with less data in image classification tasks.} A) Experiments on Fashion-MNIST image classification dataset. From left to right: 1)  Statistical efficiency: Frozen-Nets with IN-Polarity (red) always learn with fewer samples than Fluid-Net (blue); the majority of the gain is contributed by the knowledge transferred from the initial polarity configuration (pink vs. blue); Frozen-Net RAND-Polarity (green) never performs worse than Fluid-Net (blue). 2) Frozen-Net IN-Polarity always has a higher chance of reaching $80\%$ validation accuracy than Fluid-Net; Frozen-Net RAND-Polarity has comparable and sometimes a higher chance of reaching $80\%$ validation accuracy than Fluid-Net. 3) Computational efficiency: Frozen-Net IN-Polarity always takes less time than Fluid-Net to reach $80\%$ validation accuracy; again, effective knowledge transfer from the preset polarity pattern is the major contributing factor (pink vs. blue); Frozen-Net RAND-Polarity takes a similar number of computational iterations as Fluid-Net. B) Same as A except experiments were on the CIFAR-10 dataset, and the validation accuracy threshold is at $50\%$. Gray lines in the first column correspond to the validation accuracy thresholds used to plot the next two columns. For a comprehensive view of performance at different thresholds, see Supplementary Figure~\ref{fig:sup_cv}. For statistical significance of the difference, see Figure~\ref{fig:fientune}. Both datasets: n=20 trials, 100 epochs, lr=0.001. No data augmentation was performed. }
\label{fig:cv}
\end{figure}

In this section, we extend the experiments in Figure~\ref{fig:1} to image classification tasks (Figure~\ref{fig:cv}). Such complex learning tasks do not have simple and explicit polarity configuration rules as in XOR-5D. We exploited an alternative strategy by using ImageNet trained polarities (IN-Polarity). A Frozen-Net IN-Polarity has its weight \textit{magnitudes} initialized randomly, with its weight \textit{polarities} initialized and fixed to match the IN-Polarity pattern. We also tested Fluid IN-Polarity where networks were initialized to the exact same pattern as Frozen-Net IN-Polarity, except polarities are free to switch while learning the task. This comparison helps us to understand which of the two factors contributed more to the performance gain: fixing polarities or knowledge transfer through the initial polarity pattern. We trained and tested networks on the Fashion-MNIST (grayscale) and CIFAR-10 (RGB-color) datasets, using AlexNet network architecture \citep{Krizhevsky2017}. For both datasets, we trained for 100 epochs, with lr=0.001 (best out of 

[0.1,0.03,0.01,0.001]). The AlexNet IN-weights were obtained \href{https://www.cs.toronto.edu/~guerzhoy/tf_alexnet/}{here}. Of note, we controlled all conditions to follow the same weight magnitude distribution at initialization Supplementary Figure~\ref{fig:mag_dist}). Specifically, we randomly initialized the networks following conventional procedures: we used Glorot Normal for conv layers, and Glorot Uniform for fc layers; then either fixed the polarities as-is (RAND-Polarity) or flipped the polarities according to the ImageNet template (IN-Polarity), introducing no change to the magnitude distributions.

Across the board, Frozen-Net IN-polarity (red) always learn with fewer samples than Fluid-Net (blue) (Figure~\ref{fig:cv} first column). When only 100 training images were available (across the 10 classes), Frozen-Net IN-Polarity (red) yields $7\%$ less validation error at convergence compared to Fluid-Net (blue) in the Fashion-MNIST task and $9.4\%$ less error for CIFAR-10. Such a gain is mostly brought by knowledge transferred through polarity pattern (pink vs. blue, $6\%$ gain for Fashion-MNIST; $8.4\%$ gain for CIFAR-10). Fixing the polarities can further bring performance gain: for CIFAR-10, $1\%$ gain at 100 training samples, and up to $3\%$ at 50000 training samples. The improvement on validation accuracy (less validation error) from Fluid IN-Polarity to Frozen-Net IN-Polarity is often statistically significant (Figure~\ref{fig:fientune}, column 1, pink line). When the polarities are fixed randomly (RAND-Polarity, green), networks never perform worse than Fluid-Net (blue). 

Across the board, Frozen-Net IN-Polarity always learns with less time than Fluid-Net (Figure~\ref{fig:cv} third column). The majority of such a gain is brought by polarity pattern knowledge transfer. Frozen-Net RAND-Polarity takes a comparable number of computational iterations as Fluid-Net. Furthermore, not every network is able to reach the specified accuracy threshold (Fashion-MNIST: $80\%$, CIFAR-10: $50\%$; Figure~\ref{fig:cv} second column). Across the board, Frozen-Net IN-Polarity has a higher chance of passing the specified accuracy threshold than Fluid-Net; Frozen-Net RAND-Polarity has an equal and sometimes higher chance than Fluid-Net (Fashion-MNIST 1000 \& 2000 samples; CIFAR-10 10000 \& 25000 samples). These observations are true across different validation accuracy thresholds (Supplementary Figure~\ref{fig:sup_cv}). 

IN-Polarity-initialized networks in general show more consistent performance across trials compared to the Fluid setting for both Frozen-Net and Fluid-Net. This is especially obvious for the statistical efficiency plots: across the board, IN-Polarity always shows less variation in its performance on validation error (shaded area marks 25\textsuperscript{th}-75\textsuperscript{th} percentiles). 

To provide a lens into the dynamics of polarity switching throughout learning, we analyzed the ratio of weight parameters (excluding bias terms) that switched polarity between two epochs for Fluid RAND-Polarity for the first 50 epochs (first half of training) - indeed, there are more polarity switches early on in training, and the ratio decays throughout training (Figure~\ref{fig:flip}). Such a trend is true across layers and across training sample sizes. This suggests polarities are mostly learned early on during training but also remain dynamic throughout the learning process. 

Taken together, Frozen-Net IN-polarity consistently learns with less data and time and does so with a higher probability of success compared to Fluid-Net; the majority of the performance gain is brought by knowledge embedded in the initialized polarity pattern, with further gain possible by fixing weight polarities; Frozen-Net RAND-Polarity perform as well as Fluid-Net RAND-Polarity, sometimes better. 

\section{Transferring and fixing polarity is superior to transferring weights}\label{sec:finetune}

From a transfer learning perspective, Frozen-Net IN-Polarity essentially transfers weight polarities instead of weights per se. How does polarity transfer compare to the traditional finetune (weight transfer) strategy? This time, instead of randomly initializing the weight magnitudes, we initialized the weight magnitudes based on the ImageNet-trained weights (IN-Weight). We compared them with Frozen-Net IN-Polarity by plotting their differences ($\Delta$=\{LEGEND\} - \{Freeze IN-Polarity\}) in Figure~\ref{fig:fientune}. The original curves before taking the differences can be found in Supplementary Figure~\ref{fig:sup_fintune_all}.

The orange curves compare weight transfer (Fluid-Net IN-Weight) with our polarity transfer strategy (Frozen-Net IN-Polarity). Across almost all data scarcity levels, our polarity transfer strategy achieves lower validation error (higher accuracy) than finetune (first column, orange$\geqslant0$). Such a superiority is statistically significant (Mann-Whitney U two-tail; multiple comparisons corrected with Holm–Bonferroni method) when training data is limited (Fashion-MNIST: $250, 1000$ samples, CIFAR-10: $\leqslant1000$ samples except $250$). Polarity transfer also allows the networks to learn with higher probability (second column) and fewer epochs (third column, curve$\geqslant0$). Such faster learning occurs regardless of the performance threshold value (Supplementary Figure~\ref{fig:sup_finetune_test}). In sum, transferring and fixing polarity is almost always superior to weight transfer in terms of both statistical and computational efficiency. Such an observation suggests polarity configuration is an effective medium, if not superior to weight pattern, for transferring knowledge between networks. 

\begin{figure}[!ht]
\includegraphics[width=\linewidth]{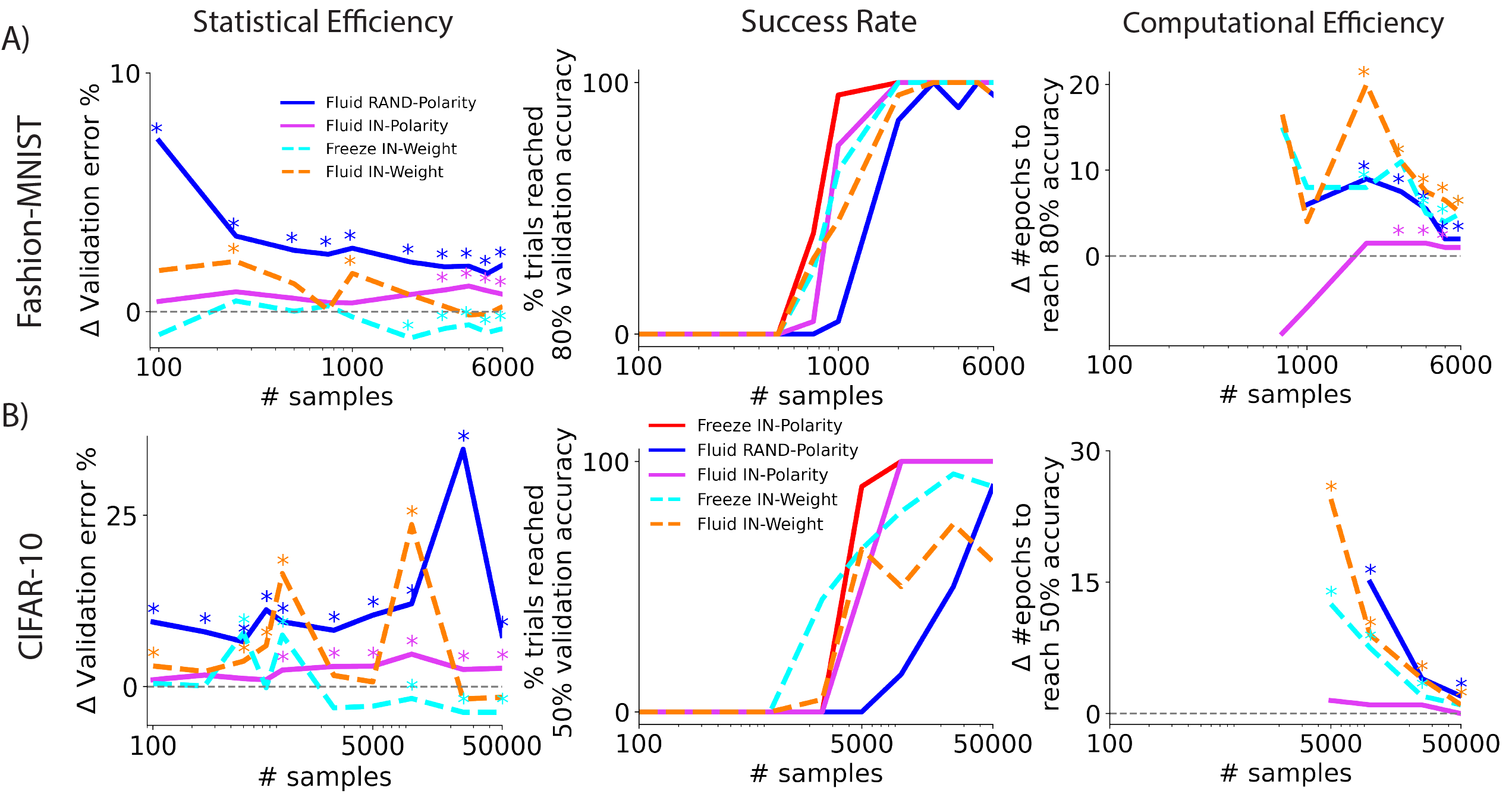}
\caption{\textbf{Across all sample sizes, Frozen-Net with ImageNet-Polarities (IN-Polarities) learns more quickly than Fluid-Net with ImageNet-Weight (IN-Weight) initialization.} In the first and third columns, curves are the \textbf{median difference} $\mathbf{\Delta}$=\textbf{\{LEGEND\} - \{Freeze IN-Polarity\}} (the second term is the red curve in Figure~\ref{fig:cv}). Asterisks (*) indicate that the validation error is significantly higher than Freeze IN-Polarity (Mann-Whitney U two-tailed test, $\alpha$=0.05). Multiple comparisons are corrected with Holm–Bonferroni method. 1) First column: Transferring and fixing polarity is more effective than transferring weights in terms of data efficiency. This is indicated by the orange curve above zero across most sample sizes, meaning Fluid-Net IN-Weight on average makes more validation error than Freeze IN-Polarity, and such differences are statistically significant for small sample sizes. Note the zero line here means LEGEND = Freeze IN-Polarity. 2) Second column: Frozen-Net IN-Polarity has a higher chance of reaching $80\%$ (Fashion-MNIST) / $50\%$ (CIFAR-10) validation accuracy compared to weight transfer (orange curve); 3) Third column: Frozen-Net IN-Polarity always takes less epochs to reach high validation accuracy than weight transfer (orange curve). Both datasets: n=20 trials, 100 epochs, lr=0.001. For the right two columns, the validation accuracy thresholds are the same as in Figure~\ref{fig:cv}.}
\label{fig:fientune}
\end{figure}

When networks were transferred with polarities but not frozen (Fluid IN-Polarity pink), they almost always perform better than Fluid IN-Weight (orange, Figure~\ref{fig:fientune} \& Supplementary Figure~\ref{fig:sup_fintune_all} to see variance), this is true except in rare cases (e.g., CIFAR-10 2500 \& 5000 samples) where the difference is not significant due to the wide performance variation of Fluid IN-Weight. 

Surprisingly, when we initialized the Frozen-Net with IN-Weight (cyan), there is some gain in performance, but to a limited extent; in fact, it can sometimes be worse. When training data is limited (Fashion-MNIST $\leqslant1000$, CIFAR-10 $\leqslant1000$), Frozen-Net IN-Weight gained little in performance (cyan$\approx$0) and could be worse at times (CIFAR-10 500 \& 1000 samples). When training data is more abundant, there is a more consistent accuracy gain by initializing Frozen-Net with IN-Weight (Fashion-MNIST gain $\sim1\%$, CIFAR-10 gain $\sim4\%$). Such a gain is discounted by more training iterations (third column), and often a lower likelihood of reaching a high level of performance (second column). 

Furthermore, similar to random initialization, weight transfers tend to have wider performance variations compared to polarity transfers, for both Frozen and Fluid networks (Supplementary Figure~\ref{fig:sup_fintune_all}). The exact reason behind this observation remains to be explored; our current hypothesis is the stochasticity of sample batching: by only transferring polarity while initializing the magnitudes randomly, the learning process is more robust against such stochasticity. An alternative contributing factor that cannot be ruled out yet is the difference in initialized magnitude distributions between polarity transfer vs. weight transfer (Supplementary Figure~\ref{fig:mag_dist}). 

In sum, polarity transfer is superior to weight transfer in most of the scenarios we tested, and such a superiority is further secured by fixing the polarities throughout training. To a large extent, weight polarity alone, not weight per se, is enough to transfer knowledge between networks and across tasks. Giving the additional magnitude information to Frozen-Net can give some performance gain, but only when data and time is abundant; in all other scenarios (i.e., data-limited or time-limited), initializing Frozen-Net with stereotypical weight magnitudes could be detrimental to the learning performance.

\section{What do we lose by setting weight polarity \textit{a priori}?}\label{sec:disadv}

Intelligent agents have limited resources in 1) data, 2) time, 3) space (number of hidden units or other network size parameters), and 4) power ($\propto $time$\times$space, number of flops). In Sections~\ref{sec:adv}, \ref{sec:cv}, and \ref{sec:finetune}, we showed that fixing weight polarity helps to save on two of these resources, time and data, but with a condition – the polarity 

\begin{figure}[!ht]
\includegraphics[width=\textwidth]{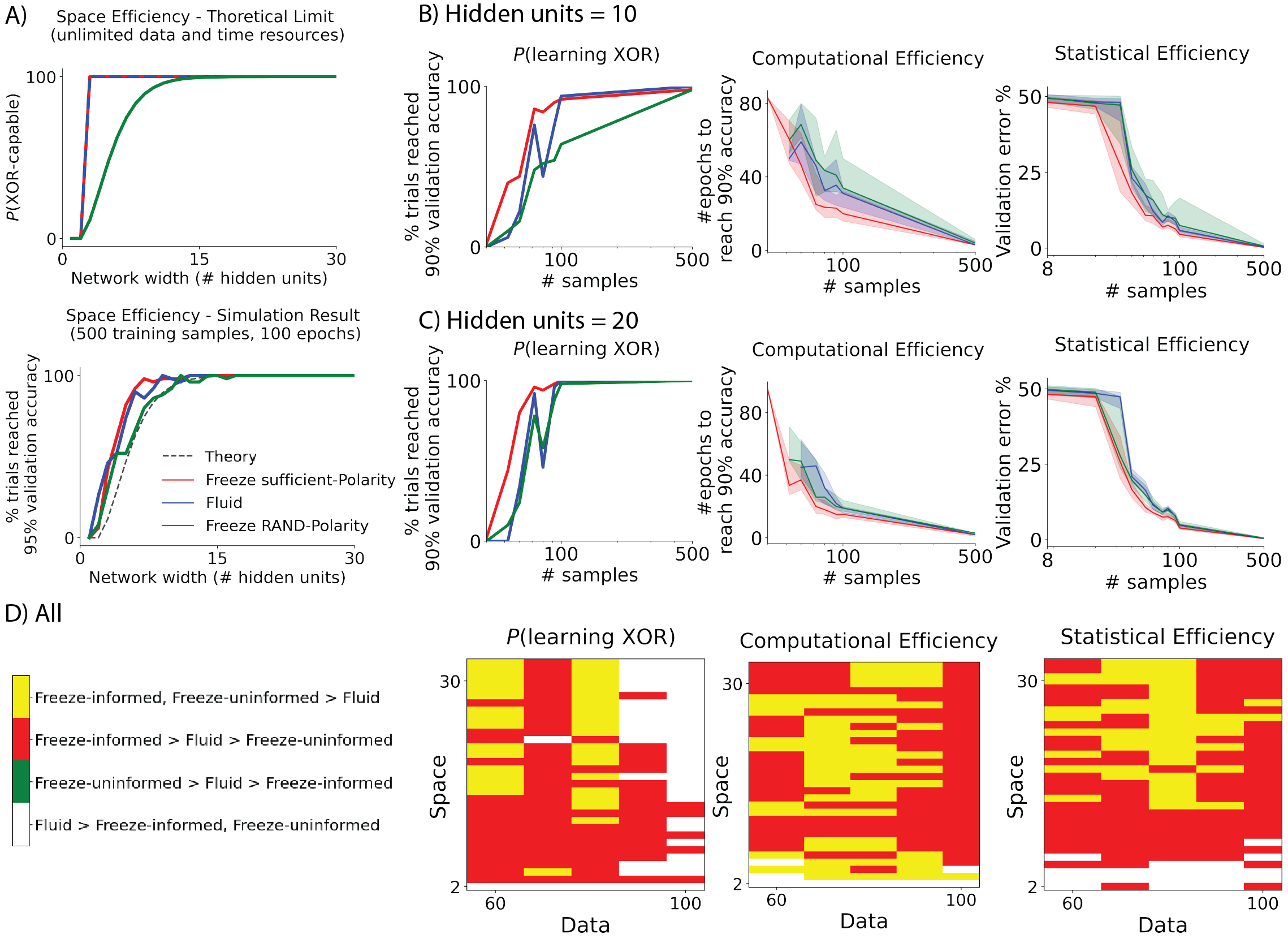}
\caption{\textbf{A network has to be sufficiently large for it to learn more quickly, with less data, if its weight polarity is configured randomly.} A) For a randomly configured Frozen-Net to learn XOR-5D, it has to be sufficiently large. In both the theoretical limit and simulation results, it takes at least 15 hidden units for a Frozen-Net RAND-Polarity (green) to have $>99\%$ chance of learning XOR-5D; it takes three hidden units, in theory, for an adequately configured Frozen-Net (red) and Fluid-Net (blue) to learn XOR-5D. B) When a Frozen-Net RAND-Polarity is not large enough, it has a much lower chance of learning XOR-5D, and when it does learn, it uses more time and data. C) When Frozen-Net RAND-Polarity is sufficiently large, it shows at least the same level of performance as standard Fluid-Nets. D) Frozen-Net Sufficient-Polarity shows an advantage over Fluid-Net (red + yellow) in almost all training sample sizes (especially when data is limited) and across all network sizes. Randomly configured Frozen-Nets have an advantage over Fluid-Nets (green + yellow), mostly when the network is sufficiently sized ($>15$ hidden units). We ran 50 trials for all experiments. All curves represent medians with shaded regions denoting 25\textsuperscript{th}-75\textsuperscript{th} percentiles. }
\label{fig:2}
\end{figure}

configuration must be adequately set such that the network can still represent the function. With fixed polarity, certain configurations will result in networks never being able to learn the task. What is the probability of such unfortunate events happening? We investigate this direction with our simulated task XOR-5D. 

Assuming unlimited resources (i.e., perfect learning algorithm, unlimited data and time), we deduced the theoretical probability limit for a single-hidden-layer network to be able to represent XOR (and its high-dimensional variants with task-irrelevant higher dimensions) as a function of the number of hidden units (Supplementary Theorem~\ref{lma7}). The theoretical results are plotted in Figure~\ref{fig:2} panel A (top). In theory, it takes at least 3 units for Fluid-Net and Frozen-Net Sufficient-Polarity to be able to learn XOR-5D on every single trial. Such a probability will never be 100\% for Frozen-Net RAND-Polarity, no matter how large the networks are. Luckily, the probability grows exponentially with the network size: having 15 hidden units is already sufficient for a randomly configured Frozen-Net to learn XOR with $>99\%$ probability. 

This is exactly what we see in our simulation results (Figure~\ref{fig:2} panel A bottom). For a generous amount of data (500 training samples) and learning time (100 epochs), the Frozen-Net RAND-Polarity curve nicely matches the theoretical curve - networks with more than 15 hidden units learn XOR-5D with very high probability, even though their polarities are fixed randomly.

Both theory and simulation results show that we will lose all advantages if weight polarities are fixed but not configured adequately; an example of such is a small, randomly configured Frozen-Net (e.g., 10 hidden units, Figure~\ref{fig:2} panel B). Notice that for the same network size, if we ensure the configuration is adequately set (red), then the network learns quickly and is data-efficient. By allowing more space (network size), Frozen-Net RAND-Polarity starts to pick up the advantages in time and data efficiency (Figure~\ref{fig:2} panel C). In summary, we gain from setting weight polarity \textit{a priori} only if the polarity is configured adequately (Figure~\ref{fig:2} panel D); the adequacy can either be made more probable by having a larger network or can be guaranteed by an explicit configuration rule (e.g., through development for NIs, or explicit rules in our simulations, or transferred polarities from previously trained networks).

\section{Discussion}\label{sec:discussion}
We showed in this paper that 1) if the weight polarities are adequately set \textit{a priori}, then networks can learn with less data and time; 2) polarity may be a more effective and compressed medium of network knowledge transfer; and 3) while we lose all advantages when the polarities are fixed but not set adequately, we can regain these advantages by increasing the size of the networks. Below we discuss the novelty of our work and some future directions. 

\paragraph{Transfer Learning} In the transfer learning literature, to the best of our knowledge, there has been no indication that transferring polarity alone is sufficient for knowledge transfer -- ours is the first demonstration on such sufficiency. Indeed, in a statistical learning framework, it is counter-intuitive to disassemble a single parameter into magnitude and sign. Previous work that vaguely took on a connectionist's view mostly focused on the existence or range of connections (e.g., adding skip connections \citep{He2015}), but the polarity of such connections were essentially left out of the discussion. Our work broke the stereotypical view and studies weight polarity as an important factor all by itself. A lesson we learned from this research is that when designing network architectures, we should not only focus on the \emph{existence} of connections, but also pay attention to the \emph{polarities} of connections.  

\paragraph{Lottery Ticket Hypothesis} Our work also agree with the results in the lottery ticket hypothesis (LTH) literature. LTH states that for a large, randomly initialized neural network, there exists small sub-networks (lottery tickets) that can be trained in isolation to match the test accuracy of the original large network \citep{Frankle2019}. It has been proven that such lottery tickets exist with high probability provided the large networks are wide enough \citep{burkholz2022on}. In our work, instead of pruning connections, we pruned half of the polarity for all connections by randomly initializing and fixing polarities. Similar to LTH theory, We also observe that larger networks enjoy a high probability of learning XOR-5D (sec~\ref{sec:disadv}). Interestingly, it has been specifically shown that polarity is important in the pruning process \citet{Zhou2019}. In essence, \citet{Zhou2019} experimentally showed that lottery tickets can no longer reach the original test accuracy if their polarity information is randomly initialized during the pruning process.  Our study agrees with them that polarity patterns contain important task-related information and should be properly initialized. Our work differs from \citet{Zhou2019} in key ways. Most fundamentally, \citet{Zhou2019} demonstrate properly presetting a subset of initialization polarity pattern is sufficient for a subnetwork to \textbf{learn at all}; we demonstrate that an adequate set of polarity patterns are sufficient for a network to \textbf{learn and transfer efficiently}. Furthermore, results in \citet{Zhou2019} can only demonstrate that polarity alone contain sufficient information to transfer knowledge across networks within \textbf{a given task}, we show polarity contain sufficient information to transfer knowledge across \textbf{different tasks} (Sec~\ref{sec:finetune}). 

\paragraph{Constrained Weight Distribution} In the computational neuroscience literature, the discussion on sign-constrained weights is done under the frame of constraining weight distributions. The discussion focuses on the memory capacity of recurrent neural networks, where the memory capacity is measured using random input-output pattern associations \citep{Amit1989, Brunel2004}. A more recent treatment of the topic of constrained weight distribution and capacity is done on a single-hidden-unit network and the network capacity is measured by packing number \citep{Zhong2022}. None of the existing work can yet relate weight distribution constraints to representation capacity of general feedforward neural networks in general task settings. This is an interesting line of work that may be useful to define the limitations of setting weight polarity \textit{a priori} in more general task settings. 

\paragraph{Bio-Plausible AI} In the literature of bio-plausible artificial neural networks (ANNs), the most related work is on Dale's principle: a single unit's output weights are exclusively excitatory or inhibitory. An exciting attempt at applying such a principle to ANNs achieved performance comparable to multi-layer perceptrons (gray-scale image classification tasks) and VGG16 (CIFAR-10) \citep{Cornford2021}. Our approach differs in several ways; the most fundamental one is ours does \emph{not} require exclusivity of a unit's weight polarity, we only ask the polarity configuration to stay fixed throughout training. Because we made fewer assumptions on the architecture and network properties, we were able to reveal the true power of weight polarities - polarity-fixed networks can not only perform as well as the traditional approaches when the polarities are set adequately, but they can also learn more quickly with smaller training samples. Additionally, we revealed that polarities, not weights, may be a more effective and compressed knowledge transfer medium. Furthermore, our Freeze-SGD Algorithm~\ref{algo} is easily applicable to any existing network architecture and any learning mode, be it transfer learning or de novo learning, thus enjoy a wider application range. 

\paragraph{Realization of the Bias-Variance Trade-off} Our work is a realization of learning theory. Our way of explicitly setting the weight polarities provides a strong prior for the networks. This is in line with the bias-variance trade-off theorem \citep{Geman1992, tibshirani2001elements}. The Frozen-Net sufficient-Polarity strategy has a high inductive bias -- by trading representation capacity, they gain in time and data efficiency through discounting parameter variances. Indeed, the performance gain we observe from fixing polarities might be explained by preventing over-fitting and thus achieving better generalization to the validation data. 

\paragraph{Future Direction: Frozen-Polarity-Compatible Learning Algorithm} We engineered our Freeze-SGD algorithm entirely based on SGD because we are interested in the pure effect of fixing polarity during learning. As discussed in Section~\ref{sec:adv}, such an approach intrinsically put Frozen-Net in a disadvantageous position by resetting weights to the correct polarity after each learning iteration, effectively fighting against the gradient update direction. As quantified in Supplementary Figure~\ref{fig:flip}, we indeed observe polarities are dynamic throughout the learning process for a Fluid network trained \textit{de novo}. It is an interesting next step to adopt a more polarity-freeze-compatible learning algorithm, possibly allowing us to further improve learning performance. One possibility is adapting the primal-dual interior-point method \citep{Vogelstein2010} or Hebbian learning \citep{Amit2019}, as well as a host of bio-plausible learning algorithms \citep{Miconi2017, Boopathy2022, Dellaferrera2022}. 

\paragraph{Future Direction: Theories and Beyond} In this work, we showed experimental evidence (simulation and image classification) that if the weight polarities are adequately set \textit{a priori}, then networks can learn with less data and time. The experimental results are interesting yet largely lack theoretical explanations. As a future direction, we are planning to work on theoretically proving polarity, compared to weights, to be a more efficient knowledge transfer medium. Our demonstration that polarity is potentially all you need to transfer knowledge across networks and tasks was done in the two settings: ImageNet to Fashion-MNIST and ImageNet to CIFAR-10. It is an important next step to test more exhaustively in different transfer settings; it is especially interesting to apply a `stress test' to the polarity transfer idea and calibrate how dissimilar two tasks have to be for polarity transfer to not work well. We also look forward to applying our Frozen-Net approach to more tasks to see if we can empirically extend our results to more diverse scenarios. This paper is a first step toward the development of supporting theory and diverse experimental tests, potentially illuminating the structure of the hypothesized "innate circuit" that enables NIs to learn rapidly.


\section*{Software and Data}
Code for running all experiments and analysis may be found on \href{https://github.com/AliceQingyangWang/weightPolarityExpr}{GitHub}. All data presented in the paper may be found \href{https://osf.io/f9wtc/?view_only=61b71c37306a41209da0eb1c35dbf8d0}{here}. 

\section*{Acknowledgements}
This work is supported by grants awarded to J.T.V:  \href{https://www.nsf.gov/awardsearch/showAward?AWD_ID=1942963&HistoricalAwards=false}{NSF CAREER Award (Grant No. 1942963)},  \href{https://www.nsf.gov/awardsearch/showAward?AWD_ID=2014862}{NSF NeuroNex Award (Grant No. 2014862)}, \href{https://www.nsf.gov/awardsearch/showAward?AWD_ID=2020312&HistoricalAwards=false}{NSF AI Institute Planning Award (Grant No. 2020312)}, and \href{https://www.nsf.gov/awardsearch/showAward?AWD_ID=2031985&HistoricalAwards=false}{NSF Collaborative Research: THEORI Networks (Grant No. 2031985)}. Q.W. was also partially supported by \href{https://reporter.nih.gov/search/uO0lXp3vF0iu1clRXTJOQg/project-details/10203331}{Johns Hopkins neuroscience T32 training grant (Grant no. 091018)}. 
The authors would like to thank Dr. Carey Priebe, Dr. Jeremias Sulam, and Dr. Konrad Kording for the helpful discussion and direction to related literatures, the \href{https://neurodata.io/about/team/}{NeuroData} lab members for helpful feedback.

\medskip
\bibliographystyle{plainnat}
\bibliography{ref}

\appendix
\clearpage
\section{Freeze-SGD algorithm}
\begin{algorithm}[H]
\caption{\textbf{Freeze-SGD}}\label{algo}
\begin{algorithmic}
    \For{$l=1,2,\ldots,L$}
        \State Get weight polarity template $T^{(l)}$ based on the configuration rules. Match $W^{(l)}$ to $T^{(l)}$
    \EndFor
    \State
    \For{$epoch=1,2,\ldots$}
        \For{$batch=1,2,\ldots$}
            \State SGD updates all weights.
    		\For{$l=1,2,\ldots,L$}
    			\State Compare signs of the weights $W^{(l)}$ to the template $T^{(l)}$, 
    			\State get $checker = T^{(l)} * \mathrm{sign}(W^{(l)})$.
    			\For{$(i,j) \text{ where } T^{(l)}(i,j)<0$}
    			\State Make $W_l(i,j)$ in compliance with $T^{(l)}(i,j)$ by one of the following four ways:
    			\State \textbf{Case posCon} $W^{(l)}(i,j) = T^{(l)}(i,j) * \epsilon$ where $\epsilon > 0$
    			\State \textbf{Case posRand} $W^{(l)}(i,j) = T^{(l)}(i,j) * rand([0, \epsilon])$ where $\epsilon > 0$
    			\State \textbf{Case zero} $W^{(l)}(i,j) = 0$
    			\State \textbf{Case flip} $W^{(l)}(i,j) = -W^{(l)}(i,j)$
    			\EndFor
    		\EndFor
    	\EndFor
    \EndFor
\end{algorithmic} 
\end{algorithm}

\newpage
\section{Methods}\label{sec:methods}
\paragraph{XOR-5D} Data were prepared by sampling from the XOR-5D distribution as described in the main text, the training data size varied, validation set is always 1000 samples across all scenarios. A single hidden layer network (64 hidden units for Figure~\ref{fig:1}, various for Figure~\ref{fig:2}) was trained for 100 epochs; in total 50 randomly seeded initializations (trials) were ran. To have the best controlled experiments, we fixed the magnitude distribution, architecture, training samples (n varies), learning rate, batch sequence of the training data, and the validation samples across all tested scenarios. 
\paragraph{Image classification} Experiments were done essentially the same as XOR-5D in a tightly controlled fashion. Across all scenarios, we controlled for the same magnitude distribution, batch sequence (batch size = 1000), learning rate, and training data. lr=0.001 was chosen from [0.1,0.03,0.01,0.001] based on the most accuracy gain after 50 epochs across all scenarios. Networks were trained for 100 epochs, 20 trials in total. We did not do any image augmentation. 
\paragraph{Statistical efficiency} \hspace{-1em} was quantified by plotting validation error rate \emph{at convergence} across different training data scarcity levels. 
\paragraph{Computational efficiency} \hspace{-1em} was quantified by plotting the number of epochs to reach certain level of validation accuracy. We note not all trails could reach the same level of validation accuracy cut-off, to present the whole picture, we 1) plotted the percentage of trials that reached the validation accuracy cut-off (\textbf{success rate}); 2) showed results across a range of cut-off selections (Supplementary Figure~\ref{fig:sup_cv}, \ref{fig:sup_fintune_all}, \ref{fig:sup_finetune_test}). 

All experiments were run on 2 RTX-8000 GPUs. 

\renewcommand\thefigure{\thesection.\arabic{figure}}

\newpage
\section{Supplementary Figures}
\setcounter{figure}{0}    
\begin{figure*}[h!]
\centering
\includegraphics[width=0.88\linewidth]{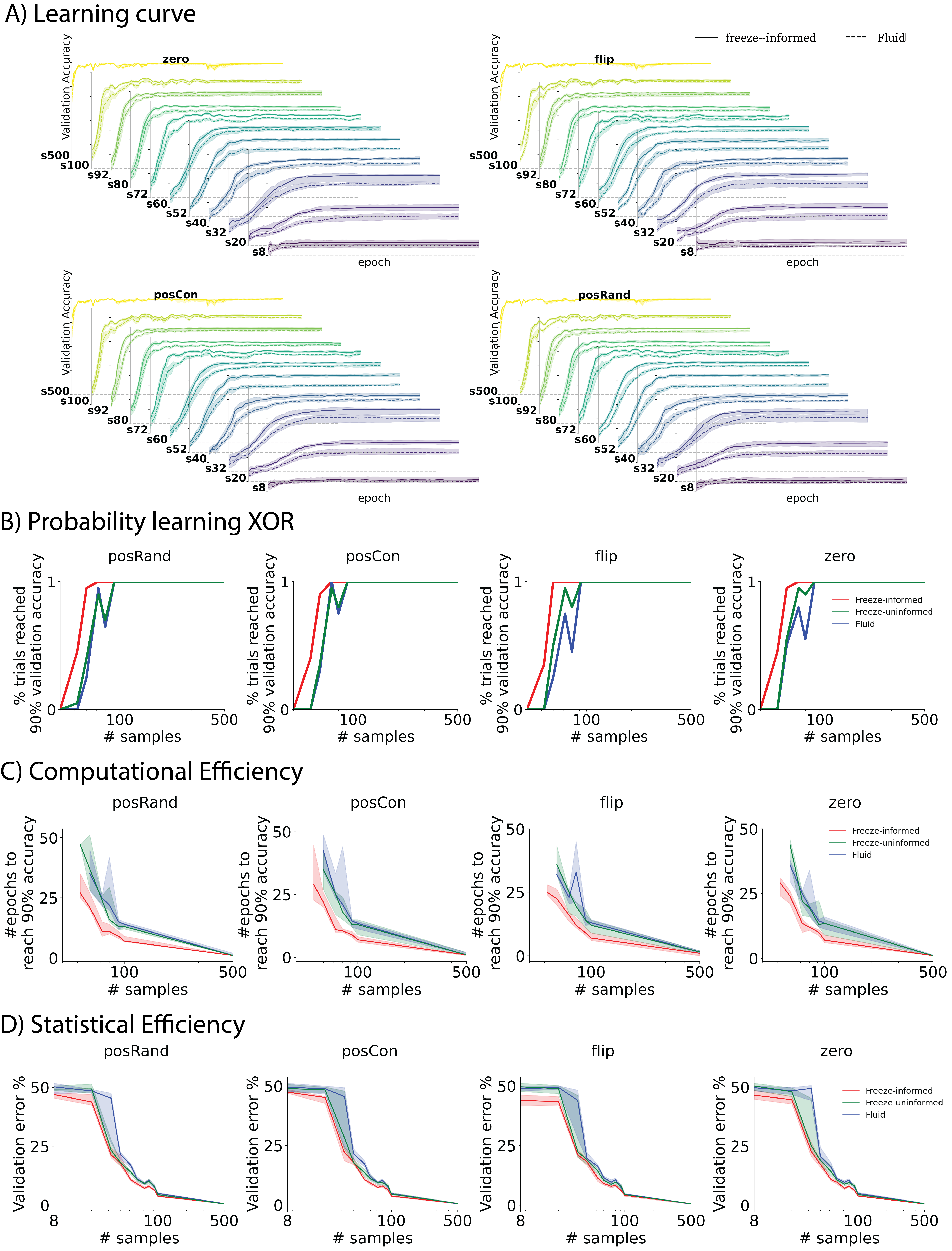}
\caption{\textbf{Different reset methods in the Freeze-SGD algorithm yields similar results.} The meaning of different reset methods is described in Algorithm~\ref{algo}. A) Freeze-informed consistently reaches higher validation accuracy after the same amount of training time. This is especially evident when training sample is scarce. B) Same curves as in Figure~\ref{fig:2}, panels B \& C first column, plotted for different reset methods. C) \& D) Same curves as in Figure~\ref{fig:1}, panels B \& C, plotted for different reset methods. We run 20 trials for all experiments in this figure. All curves correspond to medians with shaded regions representing the 25\textsuperscript{th} and 75\textsuperscript{th} percentiles.}
\label{fig:sup}
\end{figure*}

\begin{figure*}[!htbp]
\centering
\includegraphics[width=\linewidth]{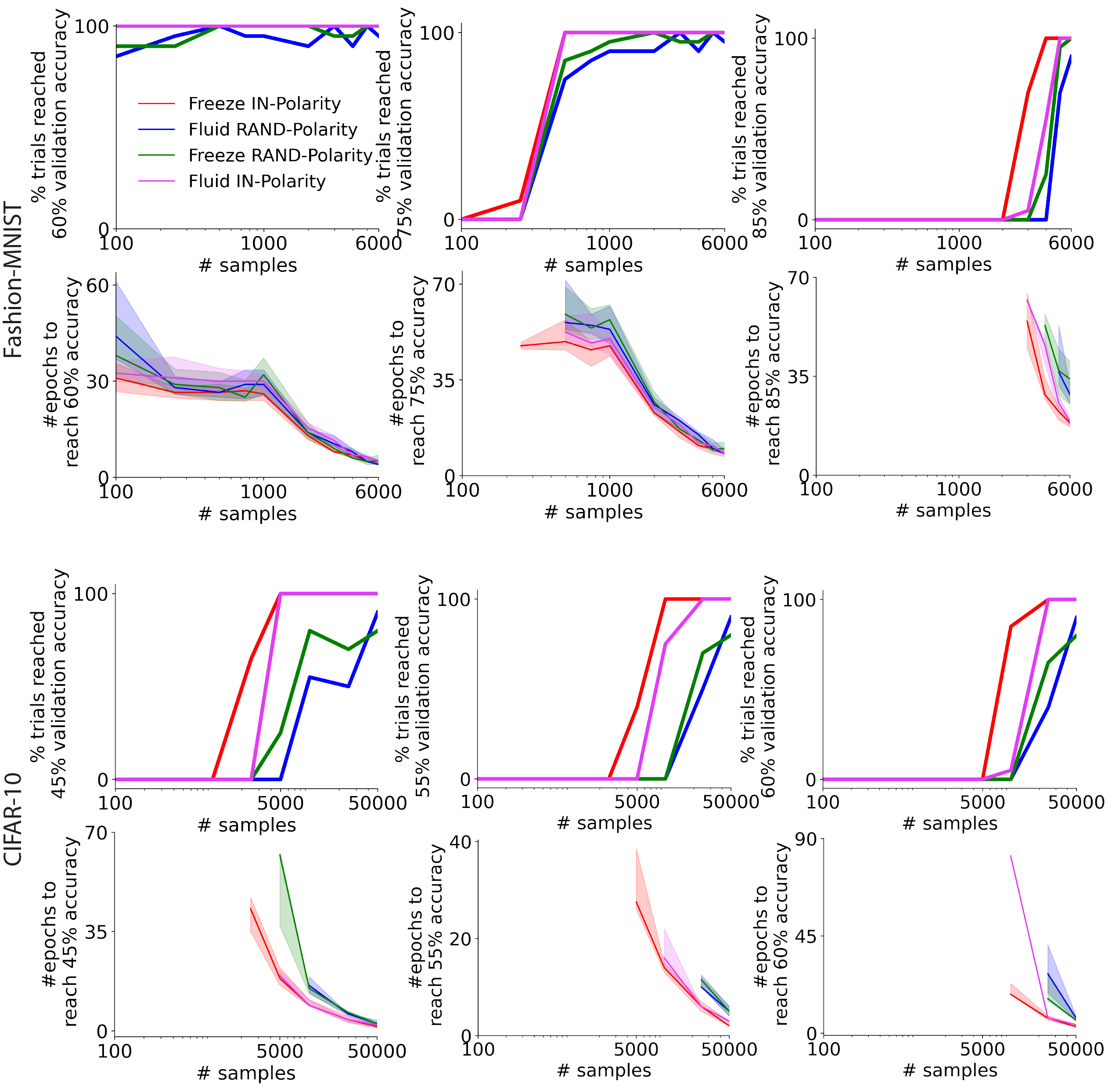}
\caption{\textbf{Related to Figure~\ref{fig:cv}. Regardless of validation accuracy threshold, Frozen-Net IN-Polarity always learn more quickly.} Same curves as in Figure~\ref{fig:cv} right two columns, plotted for different validation accuracy thresholds.}
\label{fig:sup_cv}
\end{figure*}

\begin{figure*}[!htbp]
\centering
\includegraphics[width=.9\linewidth]{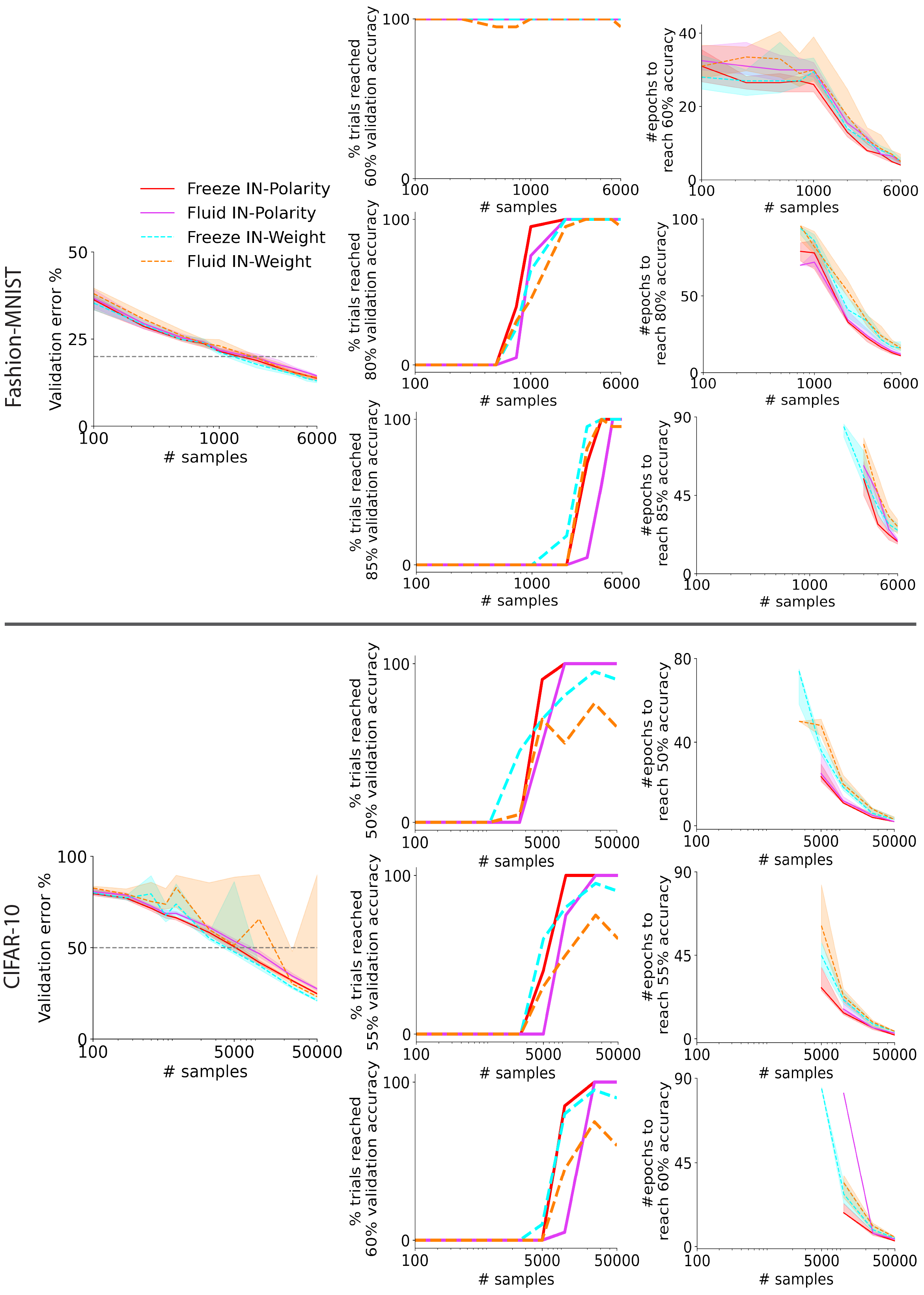}
\caption{\textbf{Related to Figure~\ref{fig:fientune}. All scenarios plotted separately instead of plotting only for the differences.}}
\label{fig:sup_fintune_all}
\end{figure*}

\begin{figure*}[!htbp]
\centering
\includegraphics[width=\linewidth]{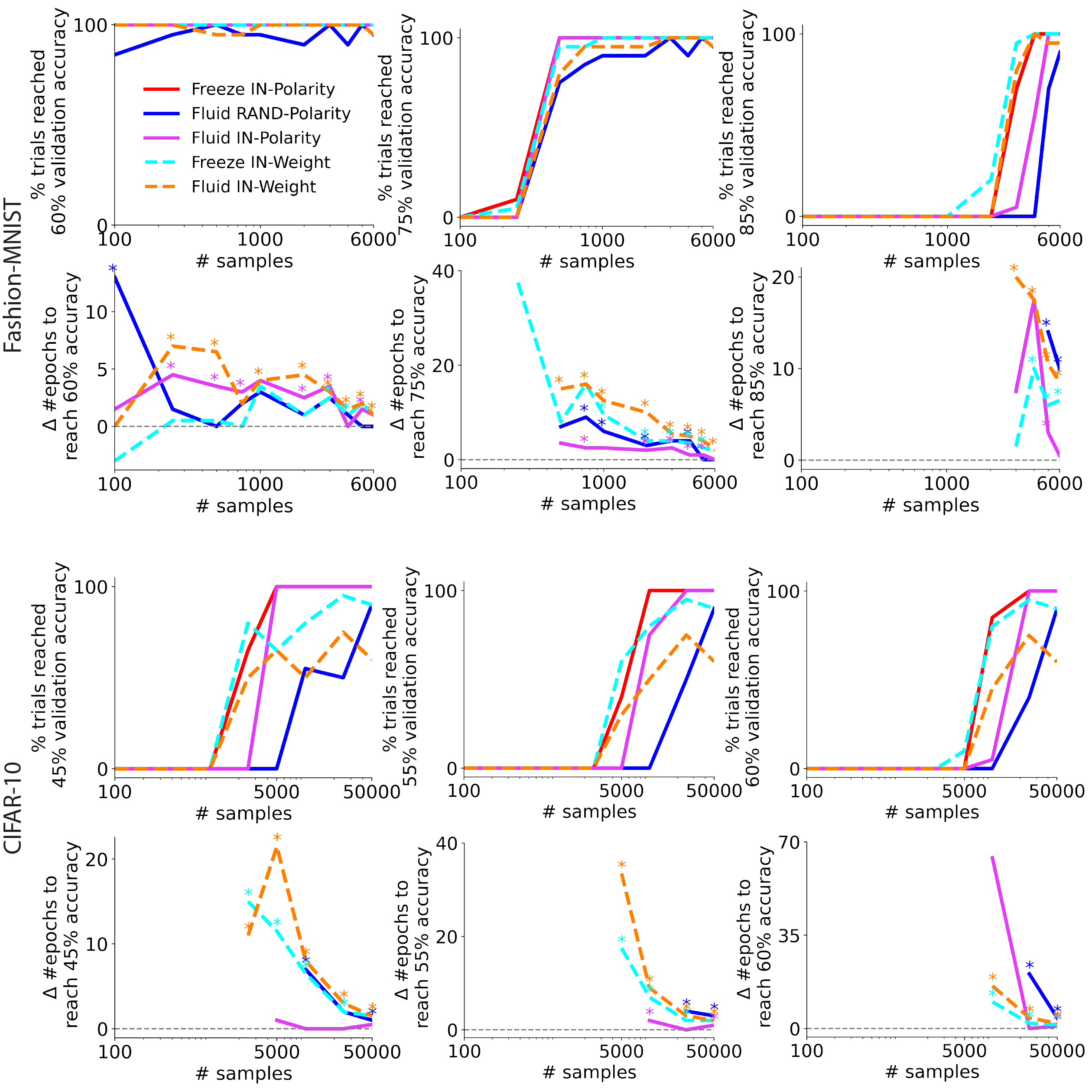}
\caption{\textbf{Related to Figure~\ref{fig:fientune}. Regardless of validation accuracy threshold, transferring and fixing polarities help networks learn faster than traditional weight transfer strategy.} Same curves as in Figure~\ref{fig:fientune} right two columns, plotted for different validation accuracy thresholds.}
\label{fig:sup_finetune_test}
\end{figure*}

\begin{figure*}[!htbp]
\centering
\includegraphics[width=\linewidth]{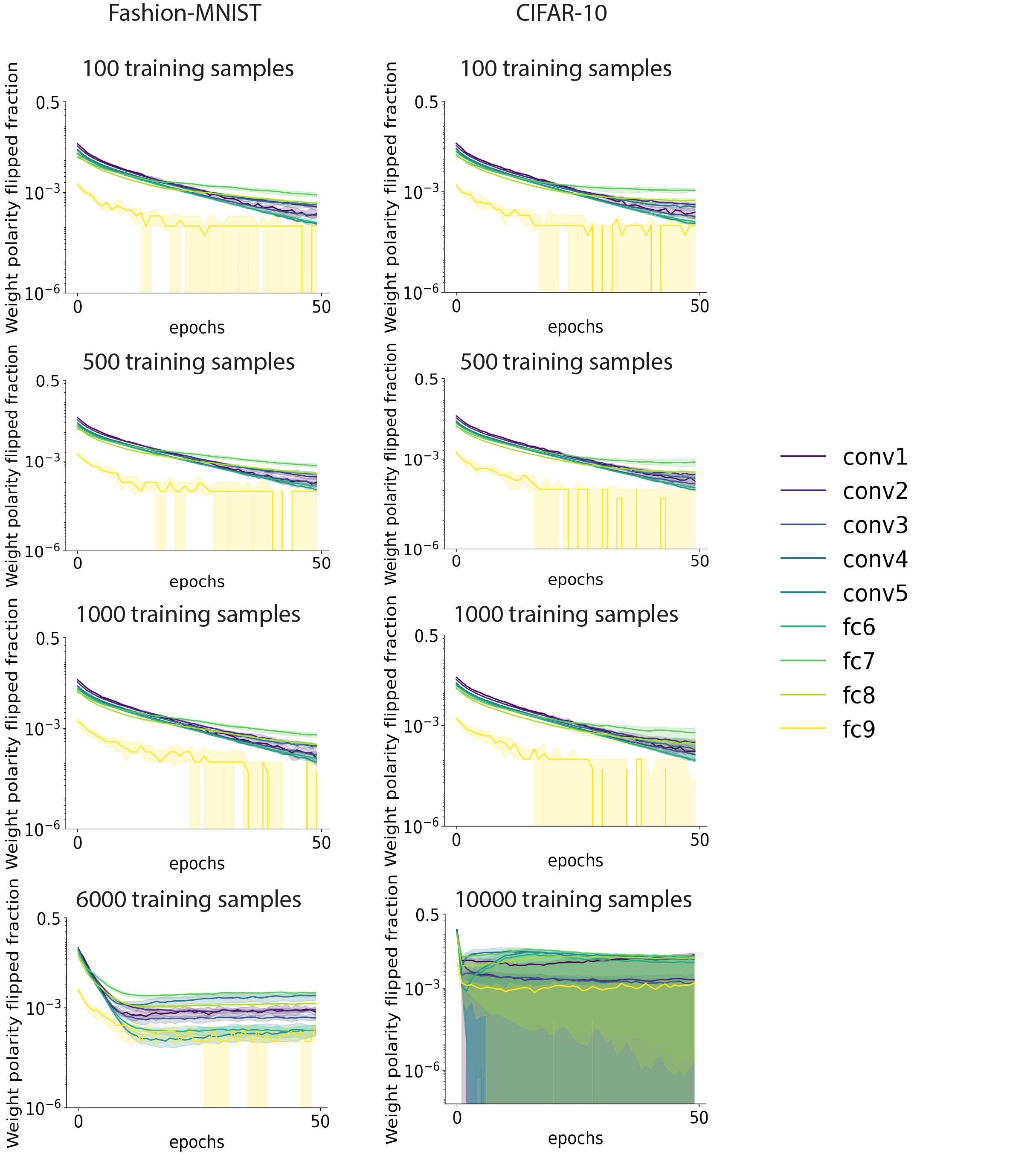}
\caption{\textbf{Tracking number of polarity flips.} Weight polarity flips were analyzed for Fluid RAND-Polarity (SGD with random initialization) and measured by the ratio of weight parameters (excluding bias terms) flipped sign between two consecutive epochs. The first 50 epochs were analyzed and plotted, separately across layers and training data size. Curves are median with shaded area representating the 25\textsuperscript{th} and 75\textsuperscript{th} percentiles out of the 20 trials.}
\label{fig:flip}
\end{figure*}

\begin{figure*}[!htbp]
\centering
\includegraphics[width=.5\linewidth]{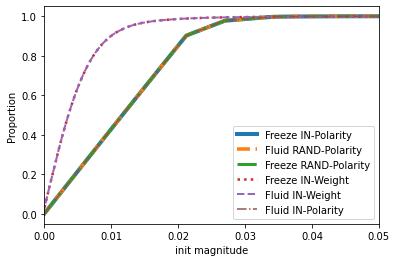}
\caption{\textbf{Weight magnitude distribution at initialization across all experimental conditions for image classification tasks.} All scenarios follow the exact same magnitude initialization magnitude except IN-weight transfer. }
\label{fig:mag_dist}
\end{figure*}

\newpage
\section{XOR Theorems related to Sec~\ref{sec:disadv}}\label{sec:supp-xor}
\begin{lemmaE}[minimum XOR solution, 2D][end,restate]
\label{lma5}
For any single-hidden-layer network to be able to solve XOR, it is sufficient to have 3 different XOR-compatible hidden units. The weight pattern for each unit can be given by a triplet $(w_{1,j}^{(1)}, w_{2,j}^{(1)}, w_{j,1}^{(2)})$, where $j$ is the unit index, $j \in \{1, \dots, n\}$, $n$ is the number of hidden units. An XOR compatible hidden unit is one where the weight pattern triplet have their polarities satisfy $sign(w_{j,1}^{(2)}) = sign(w_{1,j}^{(1)}) \times sign(w_{2,j}^{(1)})$. 
\end{lemmaE}
\begin{proofE}
For each hidden unit, we can enumerate all of its 8 possible weight polarity configurations, with the index set $A_{Polarities}=\{1,2,3,4,5,6,7,8\}$:
 
\begin{center}
\begin{tabular}{ m{1cm} m{1cm} m{1cm} m{1cm} } 
\hline
\# & $w_{1,j}^{(1)}$ & $w_{2,j}^{(1)}$ & $w_{j,1}^{(2)}$ \\ 
\hline
1 & + & + & + \\ 
\hline
2 & + & + & - \\ 
\hline
3 & + & - & + \\
\hline
4 & + & - & - \\
\hline
5 & - & + & + \\
\hline
6 & - & + & - \\
\hline
7 & - & - & + \\
\hline
8 & - & - & - \\
\hline
\end{tabular}
\end{center}

The set $B_{XOR-Polarities} = \{2,3,5,8\} \subset A_{Polarities}$ contains indexes of polarity patterns that follow XOR. To prove Lemma~\ref{lma5}, we first consider the case of 3 hidden unit single layer network. To prove a network of certain polarity pattern is capable of solving XOR, it suffices to give a working solution. Below, we exhaust all possible 3-unit network that satisfy the statement of Lemma~\ref{lma5}, i.e. having 3 different XOR-compatible units; and list out their corresponding working network solutions.


\begin{center}
\begin{table}[H]
\centering
\begin{tabular}{lllllll}
\hline
\{\} &
  \# &
  $w_{1,j}^{(1)}$ &
  $w_{2,j}^{(1)}$ &
  $b_j^{(1)}$ &
  $w_{j,1}^{(2)}$ &
  working network $F(x, y)$= \\ \hline
\multicolumn{1}{c}{\multirow{3}{*}{\{2,3,5\}}} &
  2 &
  +1 &
  +1 &
  0 &
  -1 &
  \multirow{3}{*}{$\mathrm{sigmoid}(-\sigma(x+y)+\sigma(x)+\sigma(y))$} \\ \cline{2-6}
\multicolumn{1}{c}{} &
  3 &
  1 &
  0 &
  0 &
  1 &
   \\ \cline{2-6}
\multicolumn{1}{c}{} &
  5 &
  0 &
  1 &
  0 &
  1 &
   \\ \hline
\multirow{3}{*}{\{2,3,8\}} &
  2 &
  +1 &
  0 &
  0 &
  -1 &
  \multirow{3}{*}{$\mathrm{sigmoid}(-\sigma(x)+\sigma(x-y)-\sigma(-y))$} \\ \cline{2-6}
 &
  3 &
  +1 &
  -1 &
  0 &
  +1 &
   \\ \cline{2-6}
 &
  8 &
  0 &
  -1 &
  0 &
  -1 &
   \\ \hline
\multirow{3}{*}{\{2,5,8\}} &
  2 &
  0 &
  +1 &
  0 &
  -1 &
  \multirow{3}{*}{$\mathrm{sigmoid}(-\sigma(y)+\sigma(y-x)-\sigma(-x))$} \\ \cline{2-6}
 &
  5 &
  -1 &
  +1 &
  0 &
  +1 &
   \\ \cline{2-6}
 &
  8 &
  -1 &
  0 &
  0 &
  -1 &
   \\ \hline
\multirow{3}{*}{\{3,5,8\}} &
  3 &
  0 &
  -1 &
  0 &
  +1 &
  \multirow{3}{*}{$\mathrm{sigmoid}(\sigma(-y)+\sigma(-x)-\sigma(-x-y))$} \\ \cline{2-6}
 &
  5 &
  -1 &
  0 &
  0 &
  +1 &
   \\ \cline{2-6}
 &
  8 &
  -1 &
  -1 &
  0 &
  -1 &
   \\ \hline
\end{tabular}
\end{table}
\end{center}
For the case that single-hidden-layer network has more than 3 hidden units, if it satisfies the rule in Lemma~\ref{lma5}, we can always construct a XOR-solvable network solution by setting all other weights to zero except for 3 different XOR-compatible units, then we arrive at one of the four situations in the above table and we just proved they are XOR solutions.


Therefore, for any single hidden layer network to solve XOR, it is sufficient to have at least 3 of the 4 XOR polarity patterned units. 
\end{proofE}

\begin{lemmaE}[can learn XOR probability, 2D][end, restate]
\label{lma6}
For a single hidden layer network with $n$ hidden units ($n \in \mathbb{Z}^+$), randomly initialize each weight with its polarity following $P(polarity) \sim Bernoulli(0.5)$, then the probability that this randomly initialized network can learn XOR without changing any of the weight polarity is lower bounded by
\begin{align*}
    & \Omega(P(n)) = 1-\frac{\sum_{k=0}^{2} Q(k, 4, 4, n)}{8^n}, \quad n\in\mathbb{Z}^+\\
    & Q(k, m, M, n) = [(M+k)^n-\sum_{p=0}^{k-1} Q(p, k, M, n)]\binom{m}{k}\\
\end{align*}
$Q$ here is a helper function for counting, it is defined in more detail in the proof.
\end{lemmaE}
\begin{proofE}
For any network randomly initialized, its weight polarity pattern is a set of $n$ draws with replacement from the polarities indexed by the set $A_{Polarities}$, $|A_{Polarities}|=8$. Define $H = (h_1, \dots, h_m, \dots, h_n), h_m \in A_{Polarities}, |H|= n$, to be the tuple of the indices of the observed weight patterns for the hidden layer. $h_m$ is the index of the weight polarity pattern for unit $m$. For any network to be able to solve XOR, it needs to have at least 3 units whose weight patterns are distinct members of set $B_{XOR-Polarities}$ (Lemma~\ref{lma5}). That is, let $J= \{h_m: h_m \in H, h_m \in B_{XOR-Polarities}\}$. Then we need that $|J| \geq 3$. We can define the probability of having exactly $k$ of the 4 XOR compatible weight patterns present within the hidden layer as following (for brevity, $A=A_{Polarities}, B=B_{XOR-Polarities}$):\\
None of the members in set B appears is given by:
\begin{equation*}
    P(|J| = 0) = \frac{|B|^n}{|A|^n} = (\frac{4}{8})^n = (\frac{1}{2})^n
\end{equation*}
Only one of the member in set B appeared in H, and that member can appear more than once in H is given by:
\begin{equation*}
    P(|J| = 1) = \frac{\binom{|B|}{1}((|B|+1)^n-|B|^n)}{|A|^n} = \frac{\binom{4}{1}(5^n-4^n)}{8^n}
\end{equation*}
This is explained by choosing one of 4 members of B \(\binom{|B|}{1}\), then multiply by the chosen member appears at least once $((|B|+1)^n-|B|^n)$\\
\\
Only two of the members in set B appeared in H, and both can appear more than once in H:
\begin{align*}
    P(|J| = 2) = & \frac{\binom{|B|}{2}((|B|+2)^n-\binom{2}{1}((|B|+1)^n-|B|^n)-|B|^n)}{|A|^n} \\
    = & \frac{\binom{4}{2}(6^n-\binom{2}{1}(5^n-4^n)-4^n)}{8^n}
\end{align*}
To count two members appear at least once in $H$, we have $\binom{|B|}{2}$ ways of choosing the 2 members from set $B$, and there are $(|B|+2)^n$ ways of choosing with replacement to populate the tuple $H$, where each hidden unit can choose from in total $(|B|+2)$ possible patterns, and subtract the situation where only one of the two members appeared $\binom{2}{1}((|B|+1)^n-|B|^n)$ and the situation where neither appeared $|B|^n$\\
The above equations can be put into a compact form 
\begin{equation*}
    P(|J| = k) = \frac{Q(k, |B|, |A|-|B|, n)}{|A|^n}
\end{equation*}
where $Q(k, m, M, n) = [(M+k)^n-\sum_{p=0}^{k-1} Q(p, k, M, n)]\binom{m}{k}, k\in\{0,\dots,m\}, m\in{1,\dots,|B|}, M\in{1,\dots,|A|-|B|}, n\in\mathbb{Z}^+$ is a helper function that gives exactly $k$ of the $m$ different XOR-compatible polarity patterns appeared in $n$ units, and in total $m+M$ options are considered for each unit. 

Then, $P(n)$ is counting at least 3 of the 4 set $B$ patterns appear:
\begin{align*}
        & \quad \Omega(P(n)) \\
        &= P(|J|\geq 3) \\
        &= 1-P(|J| = 0)-P(|J|=1)-P(|J|=2)\\
        &= 1-\frac{\sum_{k=0}^{2} Q(k, 4, 4, n)}{8^n}\\
        &= 1-(\frac{1}{2})^n-\frac{\binom{4}{1}(5^n-4^n)}{8^n}-\frac{\binom{4}{2}(6^n-\binom{2}{1}(5^n-4^n)-4^n)}{8^n}.
\end{align*}

\end{proofE}

\begin{theoremE}[can learn XOR probability, high dimensional][end, restate]
\label{lma7}
For a single hidden layer network with $n$ hidden units, randomly initialize each weight with its polarity following $P(polarity) \sim Bernoulli(0.5)$, then the probability that this randomly initialized network can learn high-dimensional XOR (first two dimensions are relevant, the rest $(d-2)$ dimensions are irrelevant) without changing any of the weight polarities is lower bounded by
\begin{align*}
    & \Omega(P(n, d)) = 1-\frac{\sum_{k=0}^{2} Q(k, 2^{d}, 2^{d}, n)}{(2^{d+1})^n}, \quad n, d\in\mathbb{Z}^+, d\geq2\\
    & Q(k, m, M, n) = [(M+k)^n-\sum_{p=0}^{k-1} Q(p, k, M, n)]\binom{m}{k}\\
\end{align*}
$Q$ here is a helper function for counting, it is defined in more detail in the proof.
\end{theoremE}
\begin{proofE}
We solve Theorem~\ref{lma7} with the exact same counting algorithm as in Lemma~\ref{lma6}, the only difference is now $|A_{Polarities}|=2^{d+1}$ and $|B_{XOR-Polarities}|=2^d$. We prove these two equalities below. 

$|A_{Polarities}|=2^{d+1}$ because we have $d$ input weights and $1$ output weight for each unit. 

The weight pattern of a single hidden unit in this case is given by a tuple $(w_{1,j}^{(1)}, \dots, w_{d,j}^{(1)}, w_{j,1}^{(2)}), j \in \{1, \dots, n\}$ and $n$ is the number of hidden units. Our conclusion in Lemma~\ref{lma5} trivially extends to the high dimensional case. This is because high-dimensional-XOR-solvable network solutions can be trivially constructed from Lemma~\ref{lma5} by setting the irrelevant input dimension weights to $0$. We can restate Lemma~\ref{lma5} for high-dimensional XOR as following:

For any single-hidden-layer network to be able to solve high dimensional XOR (only first two dimensions are relevant), it is sufficient to have 3 different XOR-compatible hidden units. A high dimensional XOR compatible unit polarity pattern is ruled by $sign(w_{j,1}^{(2)}) = sign(w_{1,j}^{(1)}) \times sign(w_{2,j}^{(1)})$. 

Therefore the irrelevant dimension input weights can be of either polarities and there are $2^{d-2}$ different combinations of them. Therefore $|B_{XOR-Polarities}| = 4\times2^{d-2} = 2^d$. 

Follow the steps in Lemma~\ref{lma6}, we have (for brevity, $A=A_{Polarities}, B=B_{XOR-Polarities}$)
\begin{align*}
    P(|J| = 0) = & \frac{|B|^n}{|A|^n} = (\frac{2^d}{2^{d+1}})^n\\
    P(|J| = 1) = & \frac{\binom{|B|}{1}((|B|+1)^n-|B|^n)}{|A|^n} = \frac{\binom{2^d}{1}((2^d+1)^n-(2^d)^n)}{(2^{d+1})^n}\\
    P(|J| = 2) = & \frac{\binom{|B|}{2}((|B|+2)^n-\binom{2}{1}((|B|+1)^n-|B|^n)-|B|^n)}{|A|^n}\\
    = & \frac{\binom{2^d}{2}((2^d+2)^n-\binom{2}{1}((2^d+1)^n-(2^d)^n)-(2^d)^n)}{(2^{d+1})^n}
\end{align*}
Therefore, we have 
\begin{align*}
    \Omega(P(n, d)) = & 1-\frac{\sum_{k=0}^{2} Q(k, 2^{d}, 2^{d}, n)}{(2^{d+1})^n}\\
    = & 1-(\frac{1}{2})^n-\frac{\binom{2^d}{1}((2^d+1)^n-(2^d)^n)}{(2^{d+1})^n}-\frac{\binom{2^d}{2}((2^d+2)^n-\binom{2}{1}((2^d+1)^n-(2^d)^n)-(2^d)^n)}{(2^{d+1})^n}, \\
    & n, d\in\mathbb{Z}^+, d\geq2
\end{align*}

\end{proofE}

\newpage
\section{Proofs}\label{sec:proof}
\printProofs
\end{document}